\RequirePackage[svgnames]{xcolor}

\documentclass[11pt,a4paper]{mystyle}

\usepackage[all]{hypcap}
\usepackage[svgnames]{xcolor}
\usepackage[comma,authoryear,compress]{natbib}
\renewcommand{\cite}{\citep}

\usepackage{microtype}
\usepackage{booktabs}
\usepackage{float}
\usepackage{bigstrut}
\usepackage{multirow}
\usepackage{array}
\usepackage{setspace}
\usepackage[ruled,linesnumbered]{algorithm2e}
\usepackage{ulem}

\usepackage{amsmath}
\usepackage{amssymb}
\usepackage{mathtools}
\usepackage{amsthm}
\usepackage{mathrsfs}
\usepackage{dsfont}
\usepackage{enumitem}
\usepackage{graphicx}
\usepackage{cleveref}
\usepackage{bxcoloremoji}

\usepackage{float}

\usepackage[utf8]{inputenc} %
\usepackage[T1]{fontenc}    %
\usepackage{hyperref}       %
\usepackage{url}            %
\usepackage{booktabs}       %
\usepackage{amsfonts}       %
\usepackage{nicefrac}       %
\usepackage{microtype}      %
\usepackage{amssymb}
\usepackage{fdsymbol}
\usepackage{wrapfig}
\usepackage{lipsum}
\usepackage{enumitem}
\usepackage{stackengine}
\usepackage[font=small,labelfont=bf]{caption}
\usepackage{color}
\usepackage{adjustbox}
\usepackage{soul}
\usepackage{rotating}
\usepackage{makecell}

\usepackage{xspace}
\usepackage{dsfont}
\usepackage{pifont}

\makeatletter
\DeclareRobustCommand\onedot{\futurelet\@let@token\@onedot}
\def\@onedot{\ifx\@let@token.\else.\null\fi\xspace}

\makeatother

\usepackage{soul}

\newcommand{\TASK}{MIEA}
\newcommand{\Method}{DPEC}
\newcommand{\flux}{FLUX.1-Kontext}
\newcommand{\Description}[1]{}
\let\mathcheckmark\checkmark
\renewcommand{\checkmark}{\ensuremath{\mathcheckmark}}

\title{Multi-Tool Image Editing Attribution in Facial Forgery}
\runningtitle{Multi-Tool Image Editing Attribution in Facial Forgery}

\author[1,2]{Sheng Liu}
\author[1,2]{Qiang Sheng}
\author[1,2]{Danding Wang}
\author[1]{Yu Li}
\author[1,2]{Chenming Zhou}
\author[1,2]{Juan Cao}

\affil[1]{Institute of Computing Technology, Chinese Academy of Sciences, Beijing, China}

\affil[2]{University of Chinese Academy of Sciences, Beijing, China}

\correspondingauthor={Qiang Sheng}
\emails={liusheng@ict.ac.cn, shengqiang18z@ict.ac.cn, wangdanding@ict.ac.cn, liyu@ict.ac.cn, zhouchenming21b@ict.ac.cn, caojuan@ict.ac.cn}

\begin{document}

\begin{abstract}
As generative AI tools become increasingly powerful and easy to use, people can easily edit portrait images with a prompt, necessitating the task of image editing attribution, which predicts the involved editing tools from the given image.
Existing attribution methods hold the single-tool assumption and can only attribute a specific editing tool, but struggle to handle the more complex and increasingly common multi-tool editing scenarios, where artifacts left by different editing tools are composite and overlapped.
To address this gap, we explore \textbf{M}ulti-Tool \textbf{I}mage \textbf{E}diting \textbf{A}ttribution~(\textbf{\TASK}), which aims to identify multiple editing tools involved in a multi-tool edited facial image.
To simulate the real-life editing operations on facial images, we then construct a new dataset, \textbf{MultiEdit}, which contains 500k+ edited facial images and covers six types of editing tools that support face swapping~(Deepfake) and various facial enhancements.
Inspired by the findings from data analysis, we design \textbf{\Method}, a multi-tool attribution method that can capture distinguishable, locality-aware editing tool traces from both spatial and frequency domains with the support of an error-based curriculum learning strategy. Experiments show \Method\ outperforms nine methods for facial images edited in at most five steps.

\vspace{3mm}

\coloremojicode{1F4C5} \textbf{Date}: September 2, 2026

\coloremojicode{1F3E0} \textbf{Code, Data \& Extended Materials}: \href{https://github.com/ICTMCG/MIEA}{https://github.com/ICTMCG/MIEA}

\coloremojicode{1F4AC} \textbf{Venue}: ACM Multimedia 2026

\end{abstract}

\maketitle
\vspace{3mm}

\section{Introduction}
\label{sec:intro}
With the proliferation of powerful generative AI techniques like generative adversarial networks~\cite{gan,stylegan} and diffusion models~\cite{ddpm}, the creation and manipulation of digital images has become increasingly accessible and affordable for ordinary people.
Using convenient AI image tools and software, users can produce vivid images that are hard for humans to distinguish with only a prompt~\cite{cooke2024as}, a phenomenon evidenced by its popularity on social media: 71\% of social media images are now AI-generated or AI-edited~\footnote{https://artsmart.ai/blog/ai-image-generator-market-statistics/}.

\begin{figure}[t]
    \centering
    \includegraphics[width=0.8\linewidth]{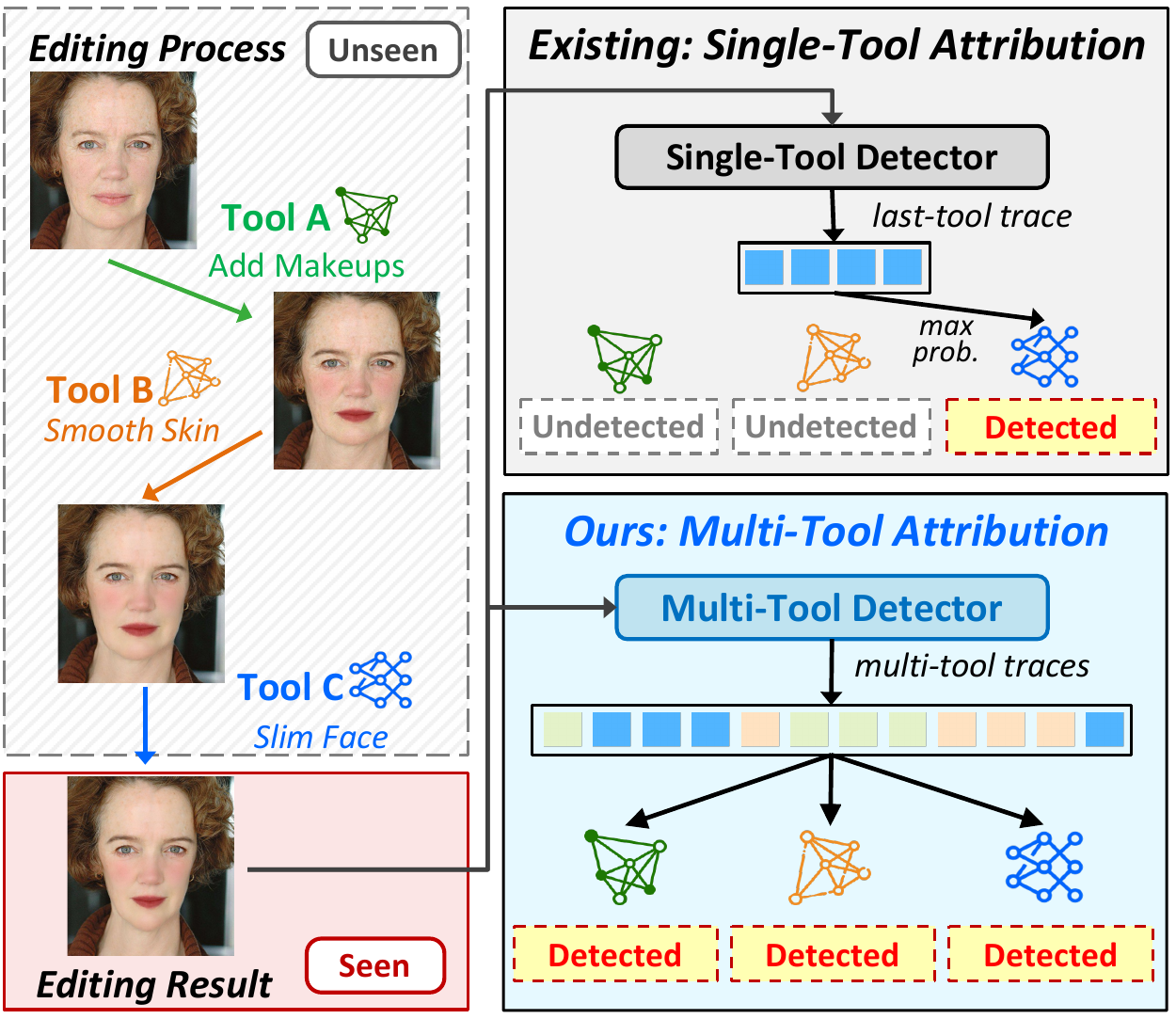}
    \caption{Multi-Tool Image Editing and Attribution. A polished portrait often undergoes multiple editing operations from different tools, and the whole editing process remains \textbf{unseen and undisclosed}. Existing single-tool attribution methods focus on extracting specific features produced by a single tool, which, to their best, only produces an incomplete prediction. In this paper, we explore how to build a \textbf{multi-tool attributor} that can identify all involved tools in the image editing process from a given edited result image.}
    \label{fig:MIEA}
    \Description{figure-1}
\end{figure}

Meanwhile, such widespread adoption of AI image tools has led to a growing risk of misuse. Fake facial images produced by AI have been used for fake news fabrication~\cite{dufour,MMFakeBench}, celebrity reputation damage~\cite{midjourney}, and academic fraud~\cite{davydiuk2025rising}, posing severe challenges to online information ecosystems.
Although researchers have made remarkable progress in AI-generated content detection~\cite{fingerprint-degrades,detection2} and attribution~\cite{DNADET,POSE}, we argue that \textbf{the recent paradigm shift from single-tool use to combinational multi-tool use in AI image editing presents a new challenge that existing solutions can hardly address}.
While some AI editing tools claim to accomplish complex edits in a single step, the final edited images circulated online are often created through multiple editing operations involving various tools~\citep{toolEdge}.
On one hand, even general-purpose image editing tools are rarely fully capable in all functions, encouraging users to apply different tools for incremental image adjustments to achieve desired effects~\citep{compEdit1,compEdit2}.
On the other hand, an image may also be edited by multiple people independently, using diverse tools for different purposes.
Consequently, multi-tool editing results in a mixture of different AI tool artifacts within a single image, violating the basic assumption that only one type of tool trace exists on an image.
This renders existing methods ineffective in practice, as they can only identify the use of one tool, and their single-tool detection sensitivity degrades significantly when other editing operations are applied afterward (we will demonstrate this empirically in Sec. 3).
If we perform forensic analysis of maliciously edited images using single-tool attribution methods, we cannot recover the full set of involved editing tools and may only detect benign edits while missing the actual malicious operations, allowing malicious actors to evade accountability. Therefore, to achieve accurate image forensics, it is critical to reconstruct the entire editing chain. By identifying every tool employed, we can demystify the black-box editing workflow, increase the transparency of the editing process, and reduce the room for hidden malicious manipulation.

To address these challenges, we carry out this work across three core dimensions: task formulation, dataset construction, and method design.
We first formulate a new task, \textbf{M}ulti-Tool \textbf{I}mage \textbf{E}diting \textbf{A}ttribution~(\textbf{MIEA}) that aims to identify multiple editing tools involved in a multi-tool edited image (Fig.~\ref{fig:MIEA}).
We then construct a large-scale facial editing dataset, \textbf{MultiEdit}, which contains over 500k edited facial images and covers six representative types of editing tools.
Based on detailed analysis of tool-specific patterns, we propose \textbf{D}ual-domain \textbf{P}atchwise \textbf{E}nhancement with \textbf{C}urriculum learning (\textbf{\Method}), a multi-tool attribution method that captures distinguishable, locality-aware editing tool traces from both spatial and frequency domains, supported by an error-based curriculum learning strategy. Moreover, \Method\ is designed to recover all tools in the image editing chain, effectively preventing early editing traces from going undetected due to being overwritten or degraded by subsequent editing operations. Experiments show that \Method\ outperforms existing methods in \TASK\ and achieves competitive performance in the conventional detection task.
In summary, our main contributions are:
\begin{itemize}[leftmargin=*, nosep]
\item \textbf{New task}: We propose a challenging task, \textbf{M}ulti-Tool \textbf{I}mage \textbf{E}diting \textbf{A}ttribution, that attributes a multi-tool-edited image to the multiple tools involved in its whole editing process, instead of only detecting the final editing tool.
\item \textbf{New method}: We propose a multi-tool attribution method  \textbf{\Method} that can extract distinguishable features of editing tools from multi-edited images via dual-domain patchwise feature enhancement and error-based curriculum learning. By uncovering the editing chain despite trace-masking from sequential edits, \Method\ enables more transparent and precise image forensics.
\item \textbf{New dataset}: We construct \textbf{MultiEdit}, a large-scale dataset covering 500k+ facial images and 6 common editing tools, on which we experimentally verify the effectiveness of \Method.
\end{itemize}

\section{Related Works}
\textbf{Face Forgery Generation \& Detection}\quad
The escalating realism of synthetic media has sparked an arms race between face forgery generation and detection techniques. Traditional editing tools like Photoshop and GIMP, which rely on manual pixel-level operations (\textit{e.g.}, splicing, liquify), often leave detectable inconsistencies~\citep{ps_detect1,ps_detect2}. With rapid advancement of deep generative models, particularly Autoencoders~\citep{ae}, Generative Adversarial Networks (GANs)~\citep{editgan,higheditgan,stylegan}, and more recently, Diffusion Models~\citep{diffedit2,diffedit} have significantly elevated the visual fidelity of various manipulation techniques, including deepfake, face reenactment, and attribute manipulation.
Accordingly, numerous deep learning-based approaches have emerged as the mainstream for manipulation detection. Specifically, Xception~\cite{xception}, TSA-Net~\citep{fasterrcnn} and constrained convolutional layers~\citep{constrainedcnn} leverage both spatial and frequency-domain features to amplify subtle manipulation traces, while $\text{F}^3\text{-Net}$~\cite{f3net} employs frequency-component partitioning for detection purposes. Recent works have further focused on cross-attention-based detectors (\textit{e.g.}, NPR~\citep{NPR} and FatFormer~\citep{FatFormer}) to cope with diverse forgery patterns in face manipulation. These efforts are supported by benchmark datasets such as ImgEdit-Bench~\citep{imgeditBench}, yet standardized evaluation protocols for multi-tool attribution still remain underdeveloped.\\
\noindent\textbf{Model Attribution}\quad
The core task of model attribution is identifying the specific tool or model behind an edit, primarily relying on proactive watermarks~\citep{watermark1,watermark2} and detecting unique model fingerprints~\cite{DNADET}. Proactive embedded fingerprints (invisible watermarks) are alternatively engineered during model training/inference for more direct provenance tracing~\citep{watermark1,watermark2}, while model fingerprint extraction uses deep learning classifiers~\citep{vixnet,POSE,FatFormer} to learn unique artifacts embedded by models like GANs. Typical contrastive learning based attribution methods, such as RepMix~\cite{RepMix} and DNA-Det~\cite{DNADET}, usually assume that an image is \textit{fingerprinted} by only one model, while multi-tool editing may mask and overwrite earlier traces, thereby rendering these methods ineffective. For example, a filter adjuster \textit{B} may destroy the traces left by the emotion-editing model \textit{A}. On the subject of multi-tool editing, Fakechain~\cite{fakechain} discovered the impacts of mixed traces on detection methods. MSA~\cite{modelship} aims to identify involved editing tools between two sequentially edited images, a special case in \TASK. Moreover, its requirement for paired sequentially edited images is too difficult to meet, as in-the-wild images hardly carry editing history records, which limits the task's practicality, while our proposed task and method aim to identify multiple editing tools used based on the final edited image only.

\section{Dataset Construction}
\subsection{Selected Editing Tools}
To investigate the traces left by different image editing tools when modifying a portrait image in a sequential manner, we selected six image editing tools based on five distinct model/algorithm frameworks: OpenCV\footnote{https://opencv.org/}, FaceFusion\footnote{https://docs.facefusion.io/}, DiffSwap~\citep{diffswap}, CSD-MT~\citep{csd-mt}, SHMT~\citep{shmt}, and \flux-[dev]~\citep{flux}. \\
\textbf{Deepfake Tools}\quad FaceFusion is an industry-leading face manipulation platform that integrates multiple advanced model architectures for face swapping, lip-syncing and so on.
DiffSwap, on the other hand, is a diffusion-based framework designed for high-fidelity and highly controllable face swapping following a conditional inpainting methodology.\\
\textbf{Face Beautify Tools}\quad OpenCV is selected as the representative non-generative method, and we implemented three types of face beauty algorithms based on its official tutorial\footnote{https://docs.opencv.org/4.x/d4/d48/tutorial\_gapi\_face\_beautification}.
CSD-MT~\citep{csd-mt} is an unsupervised makeup transfer method that decouples facial content and makeup style via frequency decomposition.
SHMT~\citep{shmt} builds upon LDM~\citep{ldm}, proposing a self-supervised hierarchical makeup transfer method and enabling precise, natural makeup transfer while preserving the source facial content. \flux-[dev]~\citep{flux} focuses exclusively on editing tasks. The model enables iterative text-guided editing, excels at character preservation across diverse scenes, and allows both precise local and global edits.

\begin{table}[t]
  \centering
  \small
  \caption{Selected tools and their editing types. Empty cell means type not supported or with unsatisfactory effects.}
  \setlength{\tabcolsep}{3.5pt}
\begin{tabular}{ccccc}
\toprule
\multirow{2}{*}{\textbf{Tool}} & \multicolumn{4}{c}{\textbf{Editing Type}} \\
\cline{2-5}
& Deepfake & Shape Beauty  & Skin Beauty & Makeup \\
\midrule
FaceFusion & \checkmark &  &  &  \\
DiffSwap   & \checkmark &  &  &  \\
SHMT       &  &  & \checkmark & \checkmark \\
CSD-MT     &  &  & \checkmark & \checkmark \\
OpenCV     &  & \checkmark & \checkmark & \checkmark \\
FLUX.1     &  & \checkmark & \checkmark & \checkmark \\
\bottomrule
\end{tabular}
\label{tab:editing_tools}
\end{table}
Considering the typical designed use scenarios of these tools and their practical editing effects, we select the editing types where these tools excel in the editing process (shown in Table~\ref{tab:editing_tools}). For each editing type, we define a set of editing targets covering frequent daily editing operations for better simulation (shown in Table~\ref{tab:editing_targets}).
\begin{table}[htbp]
  
  \small
  \centering
  \caption{Involved editing types and their editing targets}
\begin{tabular}{cl}
\toprule
\textbf{Editing Type} & \textbf{Editing Target} \\
\midrule
Deepfake & Identity change \\
\midrule
\multirow{2}{*}{Skin} & Acne/wrinkle removal \\
& Skin whitening\\
& Skin smoothing\\
\midrule
\multirow{3}{*}{Shape} & Face slim\\
& Eyes/nose/mouth size adjustment \\
& Mouth length adjustment\\
& Lips thickness adjustment\\
\midrule
\multirow{4}{*}{Makeup}
& Blush\\
& Eyeshadow\\
& Lip color adjustment \\
& Full-face makeup\\
\bottomrule
\end{tabular}
\label{tab:editing_targets}
\end{table}

\subsection{Multi-Tool Editing Workflow}

To cover as many combinations of editing tools and types as possible in real-world scenarios, we considered editing types, tools, targets, and sequential positions of tools.
\begin{algorithm}[ht]
\setstretch{0.95}
\caption{Multi-tool face editing process}\label{algorithm:plans}
\KwIn{\\
\quad Portrait set $S=\{I_1,\cdots,I_n\}$\\
\quad A portrait image $I\in S$\\
\quad Beautify steps parameter $k$\\
\quad Deepfake Tools $D=\{\text{FaceFusion, DiffSwap}\}$\\
\quad Beauty Tools\\ \quad \quad $M=\{\text{OpenCV, CSD-MT, SHMT, FLUX}\}$\\
\quad Beauty Types $E=\{\text{Shape, Skin, Makeup}\}$\\
\quad Tool Target T=$\{T_m=\{t_{m,1},\cdots,t_{m,n}\}\}_{m\in M}$\\
}
\KwResult{Edited Images $I_E=\{I_1,I_2,\cdots,I_k\}$}

$r_{\text{face}} \leftarrow \text{rand in } S\ \cup\{\text{None}\}$\;
\eIf{$r_{\text{face}} \neq {\text{None}}$}{
    $I_{\text{swap}} \leftarrow \text{randomly select from } S \setminus \{I\}$\;
    $I_1 \leftarrow \text{replace face of } I \text{ with face from } I_{\text{swap}}$\;
}
{
    $I_1 \leftarrow I$\;
}
$I_E \leftarrow [I_1]$\;
\For{$i \leftarrow 2$ \KwTo $k$}{
    $\textbf{(m,e,t)} \leftarrow \text{randomly select from } (M,E,T_m)$\;
    $I_i \leftarrow \text{apply } (\textbf{m},\textbf{e},\textbf{t})\ \text{on}\ I_{i-1}$\;
    $\text{Add}\ I_i \text{ to } I_E$\;
}
\Return $I_E$\;
\end{algorithm}

Algorithm~\ref{algorithm:plans} shows a general editing process of $k$ beautification steps.
Specifically, in line with the practical application logic of deepfake technology—where it does not perform face-swapping after facial enhancement is completed—we restrict deepfake technology to only appear in the first step of the multistep editing process.
During the actual generation process, we set $k=4$ and adopt the method of random sampling without replacement to select the editing tool $t$ used in each step. To control the data quality and evaluate the edited images, we use the SSIM~\citep{SSIM} metric to measure the similarity between the two images before and after a single editing operation, and employ Qwen3-VL~\citep{qwen-VL} to determine whether the image has achieved the editing goal. An editing process is considered successful if and only if the SSIM score falls between 0.4 and 0.98 and Qwen3-VL judges that the target editing goal has been achieved in all steps.

\subsection{Dataset Overview}
For authentic real facial images, we selected 200 real human portrait images from the DEFACTO-Face subset~\citep{defacto} as the source set S. For each image in S, we enumerate all possible permutations (i.e., unique tool-use sequences) specified in Algorithm~\ref{algorithm:plans}, and then filter out unsuccessful editing cases. Ultimately, each image is processed with 648 valid plans, yielding a total of 510k edited images in MultiEdit. The statistical details of MultiEdit are summarized in Table~\ref{tab:statistical_values}.
\begin{table}[htbp]
  \small
  \centering
  \setlength{\tabcolsep}{10pt}
  \caption{\textbf{Statistics of MultiEdit.} A permutation (Perm) is a unique tool-use editing sequence. We removed implausible permutations to better simulate real-life scenarios.}
  \begin{tabular}{c|l|r}
    \toprule
    \multicolumn{2}{c|}{\textbf{Number of Samples}} & 517,540 \\
    \midrule
    \multirow{5}{*}{\makecell{\textbf{Number of}\\ \textbf{Editing Steps}}} & 1 (w. \ \ 5 Perms)  & 43,168  \\
    & 2 (w. 17 Perms) & 129,385 \\
    & 3 (w. 36 Perms) & 129,385 \\
    & 4 (w. 48 Perms) & 129,385 \\
    & 5 (w. 24 Perms) & 86,127 \\
    \midrule
    \multirow{7}{*}{\makecell{\textbf{Number of Samples} \\ \textbf{Per Editing Tool}}}
    & FaceFusion & 172,684 \\
    & DiffSwap~\cite{diffswap} & 172,184 \\
    & CSD-MT~\cite{csd-mt} & 323,551 \\
    & SHMT~\cite{shmt} & 323,479 \\
    & FLUX.1~\cite{flux} & 323,388 \\
    & OpenCV & 323,432 \\
    \midrule
    \multirow{3}{*}{\makecell{\textbf{Number of Samples}\\ \textbf{Per Editing Type}}}
    & Shape Beauty & 86,267 \\
    & Makeup & 345,019 \\
    & Skin Beauty & 345,024 \\
    &Deepfake & 344,868\\
    \bottomrule
  \end{tabular}
  \label{tab:statistical_values}
\end{table}

\subsection{Data Analysis}
Since different editing tools may leave different traces in their editing areas, a multi-tool editing process can lead to more complex trace mixtures and overwriting issues. To guide the following method design, we perform analysis on MultiEdit samples to discover the editing impacts in single-tool detectability (Q1), domain diversity (Q2), regional effect (Q3), and inter-tool interaction (Q4).

\begin{figure}[htbp]
    \centering
    \includegraphics[width=0.7\linewidth]{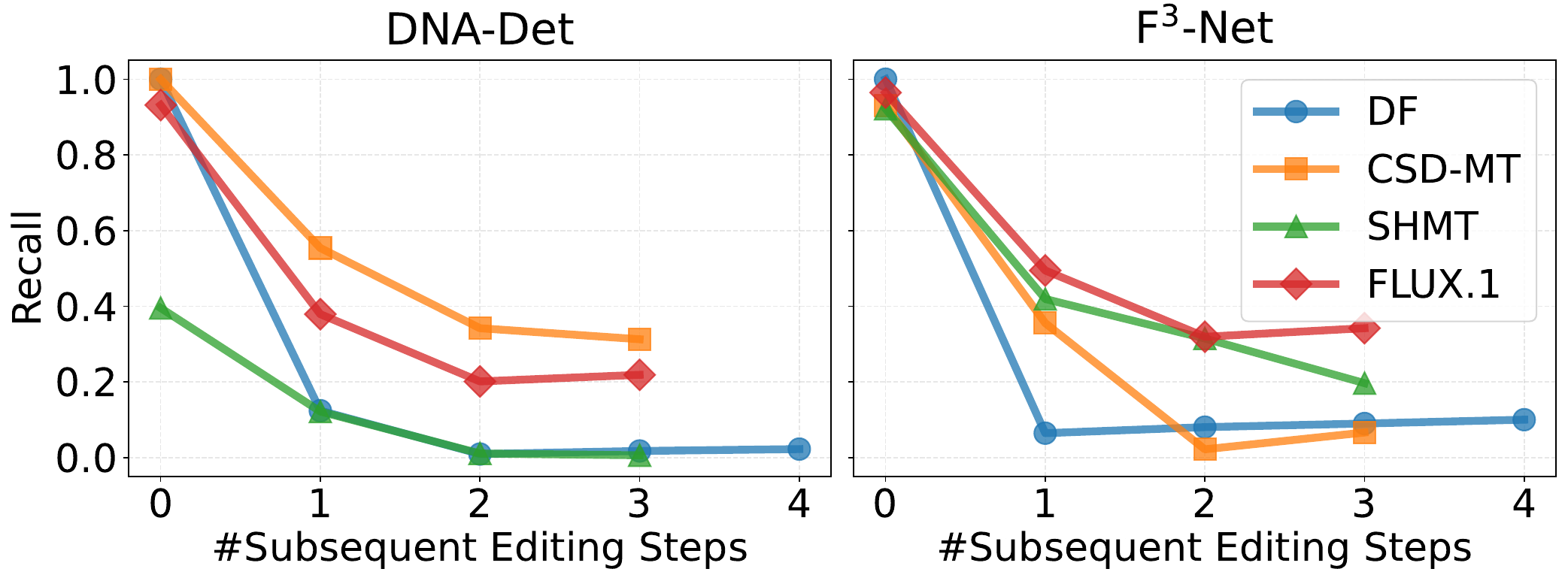}
    \caption{\textbf{Recalls of two methods ($\text{F}^3\text{-Net}$ and DNA-Det) on detecting single editing tool.} The number of subsequent editing steps indicates how many editing operations were applied after the target tool was used.}
    \label{fig:performance_decline}
    \Description{Q1-performance-degradation}
\end{figure}

\begin{figure}[ht]
  \centering
  \includegraphics[width=0.8\linewidth]{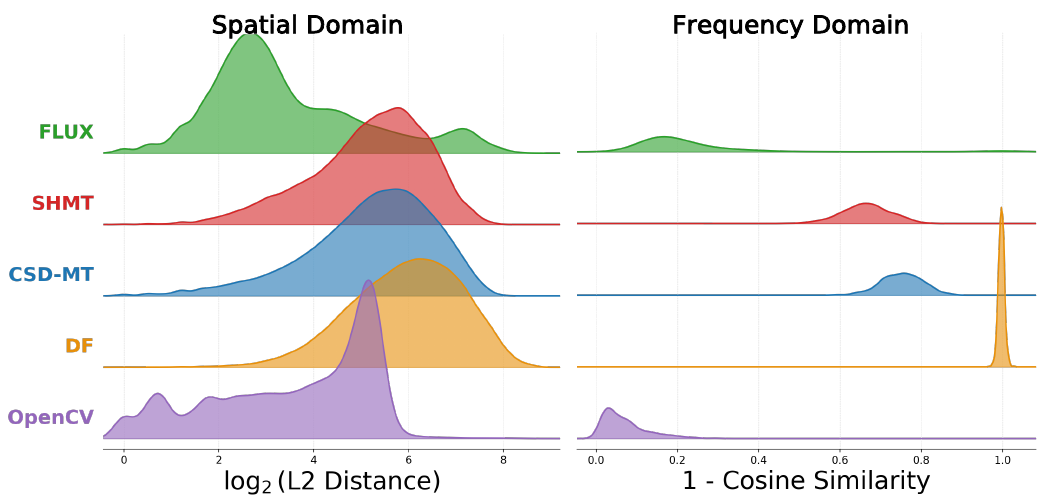}
  \caption{Ridge graph of the spatial and frequency domains among editing tools. We analyze the spatial and frequency domain differences between the images before and after editing by specific tools on different facial regions. In the spatial domain, CSD-MT, SHMT, and DeepFake show similar distributions, while totally differing in the frequency domain. Likewise, the distributions of OpenCV and FLUX are similar in the frequency domain while varying in the spatial domain.}
  \label{fig:dual-domain}
  \Description{Ridge graph of the spatial and frequency domains.}
\end{figure}

\noindent \textbf{Q1: Does multi-tool editing make tools harder to detect?}\
Previous detection methods~\citep{f3net,xception} and attribution methods~\cite{RepMix,DNADET} show good performance in detecting tools in single-tool edited images. 
To explore whether multi-tool editing makes it harder for these methods to detect specific tools, we conducted experiments where the training set comprised real images and images subjected to single-step editing solely by a specific tool, and the model was trained to perform binary classification to distinguish real from fake images. For the test set, images were produced by applying subsequent multi-step editing with other tools on the basis of the aforementioned single-step edited images, and the model was then adopted to execute the same real-versus-fake binary classification on these test images. We calculate the recall of each face beauty tool and report the mean recall across editing tools at each number of subsequent editing steps. As shown in Figure~\ref{fig:performance_decline}, as the number of subsequent editing steps increases, the recall of both detection and attribution methods shows a significant decline, indicating that \textit{\uline{the multi-tool editing indeed makes it harder to detect specific tools}}.
\newline
\noindent \textbf{Q2: What editing patterns do editing tools exhibit in spatial and frequency domains?}\ 
To investigate the patterns exhibited by different editing tools, we analyze the spatial and frequency domain similarities between the images before and after editing by specific tools on different facial regions. Let the flattened original image patch be denoted as $\mathbf{x}$ and the corresponding edited image patch as $\mathbf{y}$. For spatial domain, we compute the L2 distance by pixel-wise subtraction between the edited and the original image patches:
\begin{gather}
  D_{spatial}(\mathbf{x}, \mathbf{y}) = \|\mathbf{x} - \mathbf{y}\|_2,
\end{gather}
For frequency-domain similarity, the 2D Discrete Cosine Transform (DCT) transforms patched image data into orthogonal frequency components:
\begin{gather}
  \mathbf{X} = \text{2D-DCT}(\mathbf{x}), \quad \mathbf{Y} = \text{2D-DCT}(\mathbf{y}).
\end{gather}
Therefore, we calculate cosine similarity without the DC component to obtain the frequency-domain distance of two regions. Let $\mathbf{X}_{AC}$ and $\mathbf{Y}_{AC}$ denote the frequency vectors with the DC component removed. The cosine similarity is calculated using the dot product:
\begin{gather}
  \text{Sim}_{freq}(\mathbf{X}_{AC}, \mathbf{Y}_{AC}) = \frac{\mathbf{X}_{AC} \cdot \mathbf{Y}_{AC}}{\|\mathbf{X}_{AC}\|_2 \|\mathbf{Y}_{AC}\|_2}.
\end{gather}
Finally, the frequency-domain distance between the two regions is obtained as:
\begin{gather}
  D_{freq}(\mathbf{x}, \mathbf{y}) = 1 - \text{Sim}_{freq}(\mathbf{X}_{AC}, \mathbf{Y}_{AC}).
\end{gather}
Accordingly, we concatenate $D_{spatial}$ and $D_{freq}$ from different facial regions and plot the ridge plot to visualize both distributions for different editing tools.
As illustrated in Figure~\ref{fig:dual-domain}, the spatial domain distributions for CSD-MT, SHMT, and DF overlap considerably, yet their frequency domain profiles are clearly distinct. Similarly, OpenCV and FLUX display comparable behaviors in the frequency domain while exhibiting highly divergent distributions in the spatial domain. These findings demonstrate that editing tools sharing similar statistical patterns in one domain can still be differentiated in the other. Consequently, \textit{\uline{the spatial and frequency domains offer mutually complementary information for the \TASK\ task}}.

\begin{figure}[htbp]
    \centering
    \includegraphics[width=0.95\linewidth]{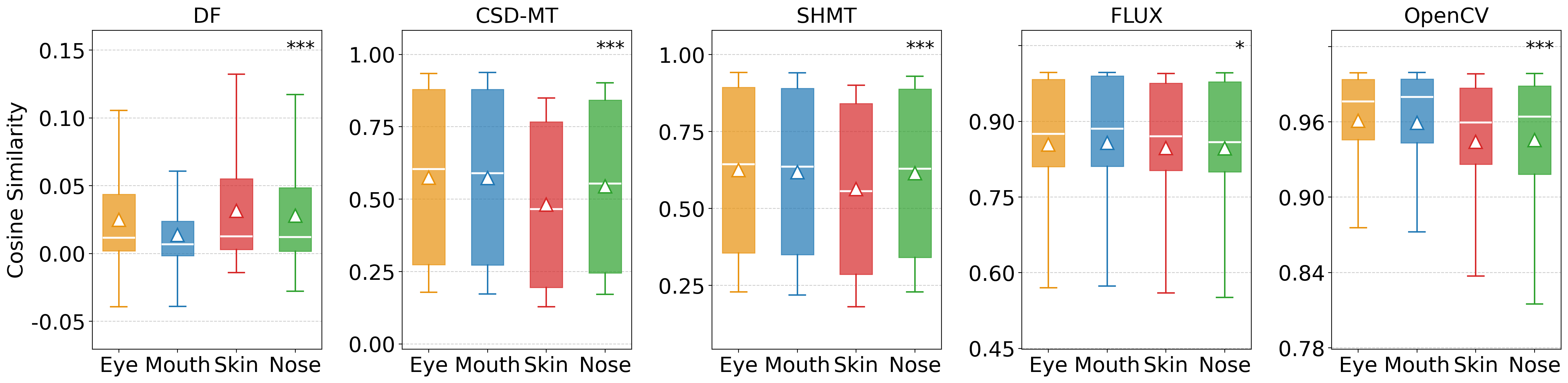}
    \caption{Grouped box plots of low and mid band of $D_{freq}$ distributions between edited and original patches across four facial regions for each editing tool. $\ast$ indicates significance levels from the Kruskal‑Wallis H test with the hypothesis that distributions among regions are not consistent.}
    \label{fig:DCT-diff}
    \Description{Grouped box plots}
    \vspace{-1em}
\end{figure}

\noindent\textbf{Q3: Does a single editing tool exhibit the same editing pattern among regions?}\ For \TASK, identifying the editing tools involved in the editing process is the core, and the key to this problem is to capture the distinguishable traces left by different editing tools.
To explore the regional characteristics of different editing tools, we conduct analysis on the frequency domain based on distribution distance to discover the consistency of editing intensity across facial regions. Inspired by~\citet{f3net} and~\citet{bandSplit}, we split the DCT coefficients into three bands (low, mid, high), select low band and mid band for further analysis by Manhattan distance, and conduct Kruskal–Wallis H test~(K-W test) to determine whether there are statistically significantly different distributions of $D_{freq}$ between edited patches and original patches among edited facial regions for each tool. Results are shown in Figure~\ref{fig:DCT-diff}. The K-W test shows that in the low and mid-frequency bands, all tools exhibit statistically significant inter-region differences, indicating that \textit{\uline{one tool can also alter different facial regions to varying degrees}}.

\noindent \textbf{Q4: What influences does the multi-tool editing process bring?}\ An intuitive thought is that as the number of editing steps increases, the edited image gradually deviates from the original image, leading to a continuous increase in human perceptual loss. From this perspective, we apply the Learned Perceptual Image Patch Similarity~(LPIPS)~\cite{lpips} metric to measure the human perceptual loss between the original image and the edited images with different numbers of editing steps. LPIPS captures high-level semantic and perceptual features to align its evaluations more closely with human visual perception, with a lower score indicating better consistency of visual features before and after editing. In Figure~\ref{fig:LPIPS}~a), from step 2 to step 5, the LPIPS score does not increase continuously, which goes against the intuition above. This counterintuitive phenomenon may be because the subsequent round of editing overwrites the relevant information of the previous round, \textit{\uline{causing the impact of the previous editing to be gradually diluted and overwritten}}.
\begingroup
\setlength{\intextsep}{6pt}
\begin{figure}[H]
    \centering
    \includegraphics[width=0.65\linewidth]{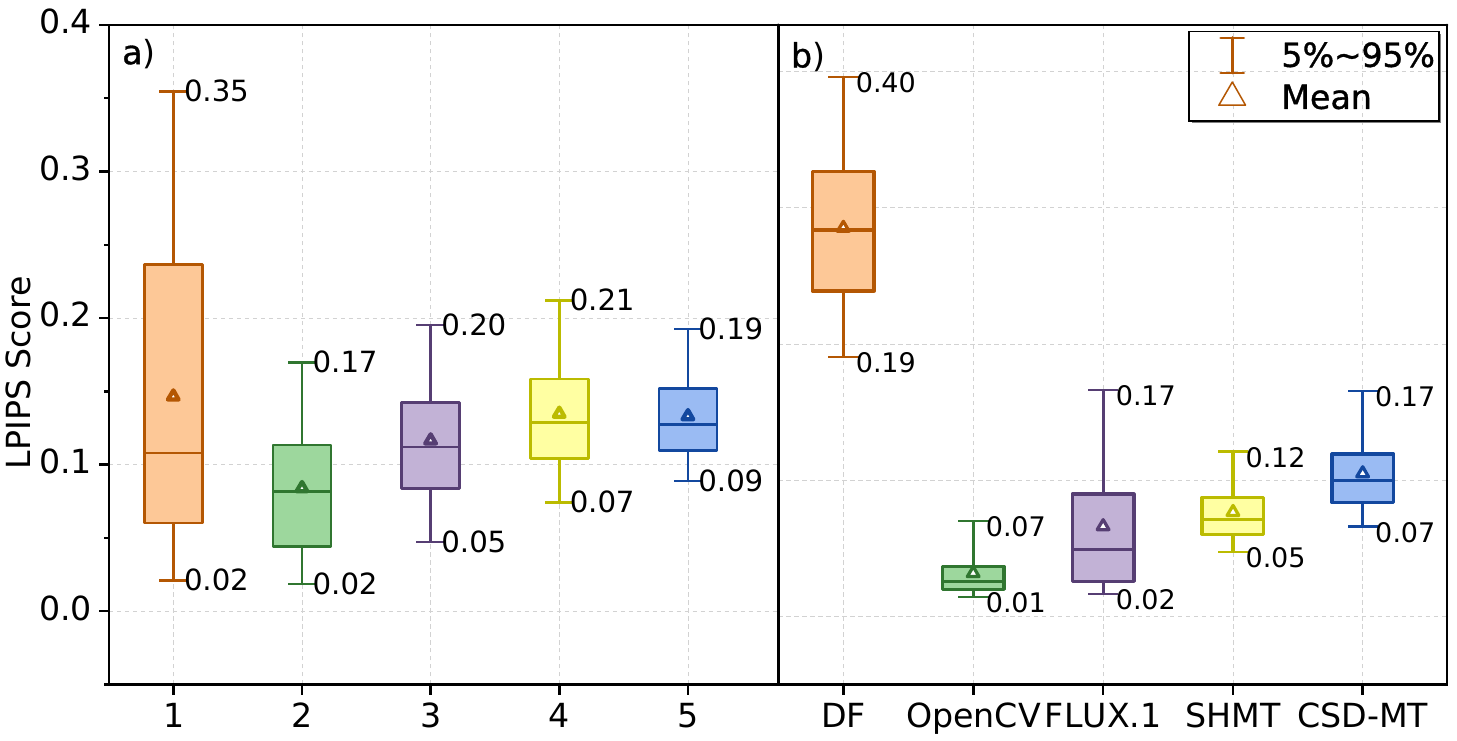}
    \caption{Box Plots of LPIPS Score. a) LPIPS Score between the original image and edited images with different numbers of editing steps; b) LPIPS Score between the images before and after editing by specific tools. DF=Deepfake Tools. Note that for results produced by deepfake tools, we calculate the LPIPS score between the target face image~(to be swapped) and the generated result.}
    \label{fig:LPIPS}
    \Description{Box Plots of LPIPS Score.}
\end{figure}
\vspace{-2em}
\endgroup

\begin{figure}[htbp]
  \centering
  \includegraphics[width=0.96\linewidth]{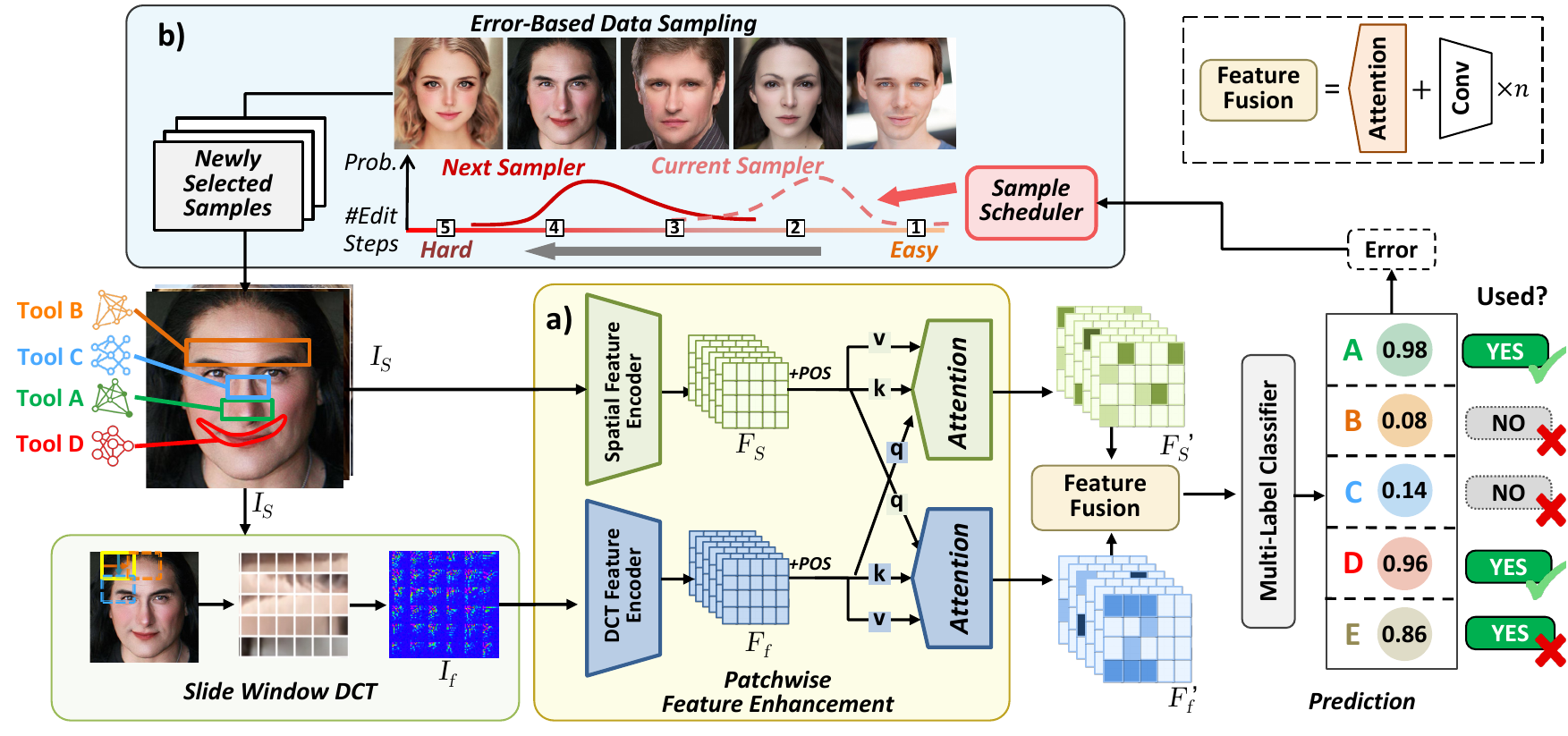}
  \caption{\textbf{Overview of \Method.} \Method\ consists of a) Patchwise Feature Enhancement (PFE) module and b) Error-based Curriculum Learning (ECL) module. PFE applies Slide Window DCT and a patchwise cross-attention to capture distinguishable, locality-aware features in both spatial and frequency domains. ECL adjusts data sampling weights of each subset of different difficulties during training based on error rate feedback, enabling the model to learn progressively from simple to hard samples. Based on the fused feature, a multi-label classifier finally predicts which tools in the candidates manipulated the given edited image $I_S$.}
  \label{fig:framework}
\end{figure}
\section{Proposed Method: \Method\ }
\subsection{Problem Formulation \& Method Overview}
We formulate multi-tool image editing attribution as a multi-label classification task: Given an edited image $I_s$, an attribution model $\mathcal{M}$ predicts which tool(s) in an $n$-tool set $\mathcal{T}$ were applied to $I_s$ editing. $\mathcal{M}$ outputs an $n$-dimensional vector for $I_s$, each entry of which corresponds to a tool in $\mathcal{T}$ and denotes the probability of its being used for $I_s$ editing.\footnote{We omit the tool-use order here for practical considerations because: 1) Detecting a tool's involvement without knowing its order is already sufficient for further forensic investigation; and 2) the feasibility of order prediction is unknown due to the possible inter-tool commutativity. The order prediction could be a future research direction.}

As presented in Figure~\ref{fig:framework}, our proposed \Method\ is a fine-grained attribution framework to figure out the involvement of editing tools in a facial image, consisting of the Patchwise Feature Enhancement~(PFE) module and the Error-based Curriculum Learning~(ECL) strategy. The PFE module is designed to extract distinguishable, locality-aware features generated by different editing tools patch by patch from both spatial and frequency domains, while the ECL strategy enables the model to progressively learn to capture individual characteristics of different tools, starting from easy samples undergoing fewer editing steps to hard samples that underwent more editing steps, by adaptively adjusting the proportion of easy and hard samples based on error feedback.
\subsection{Patchwise Feature Enhancement}
Based on the findings from Q2 and Q3 in Sec. 3.4, the PFE module is designed to select distinctive patches with the most discriminative features for different editing tools, via patch-level feature co-attention across both spatial and frequency domains. This mechanism enables position-aware patch selection, feature extraction, and dual-domain feature enhancement, all tailored to the characteristics of each editing tool, to effectively capture both intra- and inter-patch features. Formally, given an input image $I_s \in \mathbb{R}^{H\times W\times C}$, we apply Slide-Window DCT~(SWDCT)~\citep{f3net} with window size $h\times w$ and step $d$ to obtain $I_f$:
\begin{gather}
    I_{f}=\{I_{f_1},I_{f_2},\cdots,I_{f_N}\} \xleftarrow{SWDCT} I_s,
\end{gather}
where $I_{f_i} \in \mathbb{R}^{h\times w\times C}$ denotes the DCT coefficients of the $i$-th patch, and $N=(\frac{H-h}{d}+1) \times (\frac{W-w}{d}+1)$.
Then, we extract spatial features $F_s$ and frequency features $F_f$ from $I_s$ and $I_f$ using two separate backbones. Considering the different editing intensities across facial regions, we add learnable position encoding to flattened $F_s$ and $F_f$ to enhance the position awareness of features, denoted as $F_{sp}$ and $F_{fp}$, respectively.
The cross attention is computed in two directions: spatial features attending to frequency features and vice versa. Thus, the output features of PFE $F_{s}'$ and $F_{f}'$ are updated as:
\begin{gather}
  \text{Attn}(Q,K,V)=\text{softmax}\left( \frac{QK^\top}{\sqrt{d_k}} \right)V, \\
  F_{s}' =F_{s}+\text{Reshape}\left({\text{Attn}(Q_{F_{sp}}, K_{F_{fp}}, V_{F_{fp}})}\right), \\
  F_{f}' =F_{f}+\text{Reshape}\left({\text{Attn}(Q_{F_{fp}}, K_{F_{sp}}, V_{F_{sp}})}\right),
\end{gather}
where $Q_F$, $K_F$, and $V_F$ denote query, key, and value projections of $F$, and $d_k$ is the dimension of the key vectors.
\subsection{Error-Based Curriculum Learning}
Based on the observation from Q4 that simply increasing the number of editing steps will not affect the edited perceptual distance linearly, we apply an Error-Based Curriculum Learning~(ECL) strategy to progressively train the model from easy samples~(small number of editing steps with more clean traces) to hard samples~(large number of editing steps with more non-linearly mixed effects) by adjusting the proportion of easy and hard samples based on error feedback. Specifically, we first divide the training dataset into $M$ subsets according to the number of editing steps, where subset $S_i$ contains images edited with $i$ steps. During training, we maintain a dynamic sampling weight $P=\{p_1,\cdots,p_M\}$ over these subsets. At the end of each epoch, based on validation performance, we compute the error rate $e_i$ for each subset $S_i$. To adjust sampling probability, we follow the Self-Paced Learning~(SPL)~\citep{SPL} methodology and apply the polynomial soft weighting strategy~\citep{softSPL}. Formally, the training objective of the model with SPL regularizer is defined as:
{
\begin{gather}
   \min_{\mathbf{w}; \mathbf{p} \in [0,1]^\mathbb{N}} \mathbb{E}(\mathbf{w}, \mathbf{p}; \lambda) \sum_{i=1}^M p_i e_i + Q(\mathbf{p}; \lambda).
\end{gather}

}
\noindent where $\mathbf{w}$ denotes model parameters, $p_i$ is the weight assigned to the $i$-th subset (initialized to a uniform distribution), $e_i$ is the average error for subset $i$, and $\lambda$ refers to the age parameter incremented epoch by epoch.

The above learning objective is often optimized with the Alternative Optimization Strategy~\citep{SPL}. Based on the error rates $e_i$, we update the sampling probabilities $p_i$ for each subset $S_i$ as follows. Given the polynomial SPL regularizer set:
\begin{gather}
  Q = \left\{ f(\mathbf{p}; \lambda) = \lambda \left( \frac{1}{t} \sum_{i=1}^{M} p_i^t - \sum_{i=1}^{M} p_i \right) \bigg| t > 1 \right\},
  \intertext{the closed-form optimal solution for $p_i$ can be derived as:}
  p_i^* = \begin{cases}
  \left(1 - \frac{e_i}{\lambda}\right)^{1/(t-1)} & e_i < \lambda, \\
  0 & e_i \geq \lambda.
  \end{cases}
\end{gather}
$p_i^*$ decreases with respect to $e_i$ and increases with respect to $\lambda$. It holds that $\lim_{e_i \to 0} p_i^* = 1$, $\lim_{e_i \to \infty} p_i^* = 0$, $\lim_{\lambda \to 0} p_i^* = 0$, and $\lim_{\lambda \to \infty} p_i^* = 1$. That is, as the accuracy increases and the training progresses, the ECL strategy will increasingly add hard samples into subsequent training.

\section{Experiments}
\subsection{Experimental Settings}
\noindent \textbf{Evaluation Metrics}\quad We employ Strict Accuracy~(Acc-S) and Toolwise Accuracy~(Acc-T) to respectively reflect the sample and tool-level performance. Acc-S measures the accuracy of predicting the exact tools in the image editing process, while Acc-T shows the proportion of correctly identified editing tools. For example, given an image edited by three tools, if the prediction$=\{1,0,1\}$ and label$=\{1,1,1\}$, Acc-S will be 0 and Acc-T will be 0.67.

\noindent \textbf{Dataset}\quad We split MultiEdit into training and test sets in a 9:1 ratio. To avoid potential face ID bias, the original faces of training and testing samples do not overlap.\\

\noindent \textbf{Baselines}\quad Since no existing models are perfectly tailored for our task, we selected models with widely-used architectures and state-of-the-art methods for image forgery detection and attribution as baselines, including ResNet18~\citep{resnet}, ResNet50~\citep{resnet}, Xception~\citep{xception}, NPR~\citep{NPR}, $\text{F}^3\text{-Net}$~\citep{f3net}, SwinT~\citep{swinT}, and FatFormer~\citep{FatFormer}. Additionally, we employ vanilla Qwen3-VL-30B~\citep{qwen-VL} for zero-shot inference, and replace its unembedding layer with an MLP classifier (dubbed Qwen3-VL-CLS), so as to evaluate the performance of large vision-language models on this task.\\
\noindent \textbf{Implementation Details}\quad For PFE, we utilize the first three layers of Xception~\citep{xception} as the backbone for dual-domain feature extraction. The SWDCT window size is set to $16\times 16$ with a step size of 2. For ECL, we set the number of subsets $M=5$ corresponding to images with 1 to 5 editing steps, and use a polynomial SPL regularizer with $t=3$. For ECL, the initial sampling weight for each subset is set to 0.2 and the model age parameter $\lambda$ is initialized to 0.5 and increased by 1 after each epoch.
Further details are provided in the supplementary materials.

\subsection{Main Results}
\begin{table}[!t]
  \centering
  \small
  \setlength{\tabcolsep}{4.8pt}
  \caption{Performance of models at different editing steps. Acc-S=Strict Accuracy. Acc-T=Toolwise Accuracy.}
  \label{tab:main}
  \resizebox{\linewidth}{!}{
  \begin{tabular}{c|cccccccccc|cc}
    \toprule
    \textbf{\#Editing Step(s)} & \multicolumn{2}{c}{\textbf{1}} & \multicolumn{2}{c}{\textbf{2}} & \multicolumn{2}{c}{\textbf{3}} & \multicolumn{2}{c}{\textbf{4}} & \multicolumn{2}{c}{\textbf{5}} &\multicolumn{2}{|c}{\textbf{Mean}}\\
    \midrule
    \textbf{Method} & Acc-S & Acc-T & Acc-S & Acc-T & Acc-S & Acc-T & Acc-S & Acc-T & Acc-S & Acc-T & Acc-S & Acc-T\\
    \midrule
    ResNet18\cite{resnet} & 0.7982	&0.9557	&0.7123	&0.9344	&0.5271	&0.8944	&0.4096	&0.8590	&0.3485	&0.8444 &0.5372	&0.8924\\
    ResNet50\cite{resnet} & 0.8889 & 0.9766 & 0.8647 & 0.9715 & 0.7190 & 0.9416 & 0.5967 & 0.9116 & 0.5249 & 0.8941&0.7070&0.9366 \\
    Xception\cite{xception} & 0.8513 & 0.9689 & 0.7008 & 0.9372 & 0.5435 & 0.8995 & 0.3611 & 0.8409 & 0.2567 & 0.8008&0.5153&0.9061 \\
    SwinT\cite{swinT} & 0.9444 & 0.9878 & 0.8308 & 0.9646 & 0.6059 & 0.9190 & 0.4224 & 0.8778 & 0.3161 & 0.8537&0.5965&0.9150 \\
    $\text{F}^3\text{-Net}$\cite{f3net} & 0.9753 & 0.9950 & 0.9524 & 0.9904 & 0.9131 & 0.9826 & 0.8550 & 0.9705 & 0.8094 & 0.9614 &0.8963&0.9790\\
    Qwen3-VL & 0.0498 & 0.6702 & 0.0561 & 0.5110 & 0.0005 & 0.4788 & 0.0000 & 0.4664 & 0.0000 & 0.3989&0.0184&0.4864 \\
    Qwen3-VL-CLS & 0.3990 & 0.8411 & 0.1532 & 0.7402 & 0.1668 & 0.7570 & 0.1700 & 0.7200 & 0.1764 & 0.7331&0.1852&0.7466 \\
    NPR\cite{NPR} & 0.7731 & 0.9027 & 0.3894 & 0.8537 & 0.3344 & 0.8235 & 0.2145 & 0.7705 & 0.2294 & 0.7684 &0.3250&0.8152 \\
    FatFormer\cite{FatFormer} & 0.9514 & 0.9902 & 0.9070 & 0.9812 & 0.8066 & 0.9608 & 0.7029 & 0.9395 & 0.6108 & 0.9209 &0.7853&0.9564\\
    \rowcolor{Blue!10}
    \textbf{\Method} & \textbf{0.9938} & \textbf{0.9988} & \textbf{0.9806} & \textbf{0.9961} & \textbf{0.9309} & \textbf{0.9861} & \textbf{0.8918} & \textbf{0.9779} & \textbf{0.8740} & \textbf{0.9742} &\textbf{0.9293}&\textbf{0.9856}\\
    \bottomrule
  \end{tabular}
  }
\end{table}
\textbf{\Method\ outperforms all baselines on edited images of all difficulty levels}\quad Specifically, \Method\ achieves a mean strict accuracy of 92.93\% and a mean toolwise accuracy of 98.56\%, significantly surpassing the baselines and SOTA methods. Also, its performance on the most difficult cases~(5 editing steps) is superior, with 87.40\% strict accuracy. This highlights the superiority of \Method\ in addressing the challenges posed by multi-tool image editing attribution.

Moreover, consistent with the results in~\citep{fakechain}, as the number of editing steps increases, all models suffer varying degrees of performance degradation. This is primarily because the traces left by different editing tools on the image overlap and mask each other more severely. Further, methods are tailored to extract features from single-tool edited images and attribute all observed features exclusively to a single tool, naturally restricting their capacity to analyze the complex and cumulative effects.

\subsection{Performance on Binary Detection}
Although \Method\ is designed to detect the involvement of editing tools in the post-detection stage, it can also be extended naturally to detect fake images. To evaluate the traditional detection performance, we conducted experiments where we replaced all deepfake-tool-edited images in the training set with images from FF++ and collected the prediction results associated with the deepfake tag among the multi-label prediction outcomes to determine whether the image is fake or not. As shown in Table~\ref{tab:binary_detection_cross_dataset}, \Method\ achieves performance competitive with those baselines on both FF++~(in-dataset) and Celeb-DF~(cross-dataset).
This also demonstrates that \Method\ can effectively distinguish between real and edited images, which is crucial for practical digital forensics and media authenticity.
\begin{table}[ht]
  \centering
  \small
  \setlength{\tabcolsep}{4pt}
  \caption{Cross-dataset binary detection performance. For baseline methods, we directly apply binary classification heads after their backbone networks and train the whole model on the respective datasets. For \Method, we replace all Deepfake-tool-edited images with those from the respective datasets and choose the toolwise prediction result~(Deepfake tool) to determine whether the image is fake or not.}
  \label{tab:binary_detection_cross_dataset}
  \begin{tabular}{l c cc}
    \toprule
    \multirow{2}{*}{\textbf{Model}} & \multirow{2}{*}{\textbf{Train On}} & \multicolumn{2}{c}{\textbf{Test Dataset (ACC)}} \\
    & & \textbf{FF++} & \textbf{Celeb-DF~\cite{celebdf}} \\
    \midrule
    ResNet50\cite{resnet} & \multirow{5}{*}{FF++\cite{ffpp}} & 0.887 & 0.615 \\
    Xception\cite{xception} &  & 0.919 & 0.695 \\
    $\text{F}^3\text{-Net}$\cite{f3net} &  & 0.926 & 0.788 \\
    NPR\cite{NPR} &  & 0.925 & 0.726 \\
    \rowcolor{Blue!10} \textbf{DPEC} &  & 0.922 & 0.786 \\
    \bottomrule
  \end{tabular}
\end{table}

\subsection{Performance on Compressed Images}
In real-world scenarios, the edited images may be compressed and resized when being uploaded to social media platforms, which can further obscure the traces left by editing tools. To evaluate the robustness of \Method\ under such conditions, we apply common compression and resizing operations to the test set and assess the model's performance. The results show that while there is a decline in accuracy compared to uncompressed images, \Method\ still maintains a significant performance advantage over baseline methods, demonstrating its robustness in practical applications.

\begin{figure}[H]
  \centering
  \includegraphics[width=1.00\linewidth]{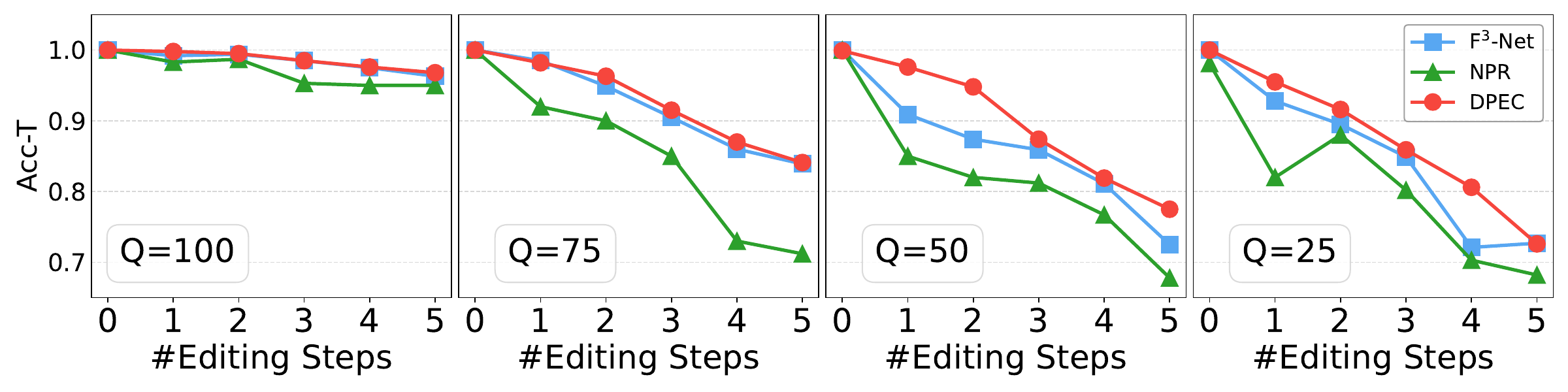}
  \caption{Performance comparison on compressed images.}
  \label{fig:compressed_performance}
  \Description{Performance comparison on compressed images.}
\end{figure}

\subsection{Ablation Study}
\noindent \textbf{Module-Level: Effectiveness of PFE and ECL}\quad To validate the effectiveness of each component in \Method, we conduct ablation studies by removing the PFE module, the ECL strategy, or both, respectively. As shown in G1 of Table~\ref{tab:ablation_study_module}, removing either PFE or ECL results in a significant drop in performance, demonstrating their crucial roles in feature extraction and progressive learning.
\begin{table}[H]
  \small
  \centering
  \setlength{\tabcolsep}{3.5pt}
  \caption{Strict accuracies~(Acc-S) of \Method\ and the ablative variants at the module (G1) and transform (G2) levels. Results with 3 to 5 editing steps are reported for brevity. The model without PFE uses the vanilla backbone for feature extraction. The patch size for image splitting is set to match the sliding window size of SWDCT.}
  \label{tab:ablation_study_module}
  \begin{tabular}{llccc}
    \toprule
    \multicolumn{2}{c}{\textbf{\#Editing Steps}}& \textbf{3} & \textbf{4} & \textbf{5}  \\
    \midrule
    \rowcolor{Blue!10} & \textbf{\Method} & \textbf{0.9309} & \textbf{0.8918} & \textbf{0.8740} \\
    \midrule
    \multirow{3}{*}{G1: Modules}  &\ \ w/o PFE      & 0.9121 & 0.8325 & 0.8234  \\
   & \ \ w/o ECL      & 0.9270 & 0.8872 & 0.8568  \\
    &\ \ w/o ECL \& PFE  & 0.8278 & 0.7968 & 0.7297 \\
    \midrule
    \multirow{2}{*}{G2: Transforms}      & Patched-DWT      & 0.8223 & 0.7857 & 0.7438  \\
    & Patched-DCT      & 0.8875 & 0.8226 & 0.8097  \\
    \bottomrule
  \end{tabular}
\end{table}

\noindent\textbf{Transform-Level: Effectiveness of SWDCT}\quad To assess the impact of SWDCT in the PFE module, we replace it with standard DCT and DWT and evaluate the model's performance.
As shown in G2 of Table~\ref{tab:ablation_study_module}, the original model that adopts SWDCT outperforms the variants by approximately 14.7\% and 7.1\%, respectively. This superiority remains consistent across different editing steps, demonstrating that the SWDCT is more effective in providing distinguishable frequency-domain features for \TASK.

\subsection{Further Analysis}
\noindent \textbf{Impact of the Size of Editing Areas}\quad To investigate whether large-area editing operations at the later steps impact the performance, we analyze the performance on the samples with editing lengths of 4 and 5. Based on whether the total editing area in the first/last two steps exceeds 50\% of the face region, we categorize them into three groups (\textit{i.e.}, large-to-small, large-to-large, and small-to-large) and calculate the respective tool attribution performance.\footnote{Due to tool use limitations, there are no small-to-small samples.} From Table~\ref{tab:subset_performance}, we see that when the large-area editing operation appears in later steps, identifying tools used earlier becomes harder. This is intuitive, as large-area editing operations are more likely to overwrite or obscure the traces left by previous editing tools, making the attribution more challenging. Despite this, our method still outperforms the best baseline $\text{F}^3\text{-Net}$ with $\text{Acc-S}=81.32\%$ on small-to-large samples.
\begin{table}[ht]
  \centering
  \small
  \setlength{\tabcolsep}{2.8pt}
   \caption{Performance on different subsets. The last row shows the largest performance drop in each column.
  }
  \begin{tabular}{ccccc}
    \toprule
    \multicolumn{2}{c}{\textbf{Size of Edited Area}} & \multirow{2}{*}{\textbf{Acc-S}} & \multirow{2}{*}{\textbf{Acc-T}} &\textbf{Recall} \\
    First 2 Steps & Last 2 Steps & & & Step 1 / 2 \\
    \midrule
    Large & Small & 0.9563 & 0.9781 & 0.9970 / 0.9593 \\
    Large & Large & 0.9437 & 0.9713 & 0.9834 / 0.9592 \\
    Small & Large & 0.8512 & 0.9256 & 0.9107 / 0.9404  \\
    \rowcolor{Blue!10}\multicolumn{2}{c}{$\Delta(\downarrow)$} &\textit{0.1051} &\textit{0.0525}& \textit{0.0873} / \textit{0.0189}\\
    \bottomrule
  \end{tabular}
  \label{tab:subset_performance}
\end{table}
\newline
\noindent\textbf{Decisive ROI Visualization}\quad To discover which facial regions play a decisive role when attributing different editing tools, we employ Grad-CAM~\citep{grad-CAM} to visualize the regions that \Method\ focuses on when identifying each editing tool. The visualizations reveal that \Method\ effectively attends to relevant facial areas corresponding to the specific edits made by each tool. Also, the ROIs are distinct for different tools, showing the effectiveness of the designed PFE module, selecting the most tool-relevant representative image patches to attribute for each editing tool. However, an interesting observation is that, for some tools like OpenCV, the model probably focuses on non-facial regions such as the neck and hair. The reason could be that the same facial skin regions are also affected by other editing tools, making it challenging to isolate the effect of a single tool solely based on facial areas. Thus, the model may leverage cues from adjacent non-facial regions to enhance attribution accuracy.
\begin{figure}[H]
  \centering
  \includegraphics[width=0.80\linewidth]{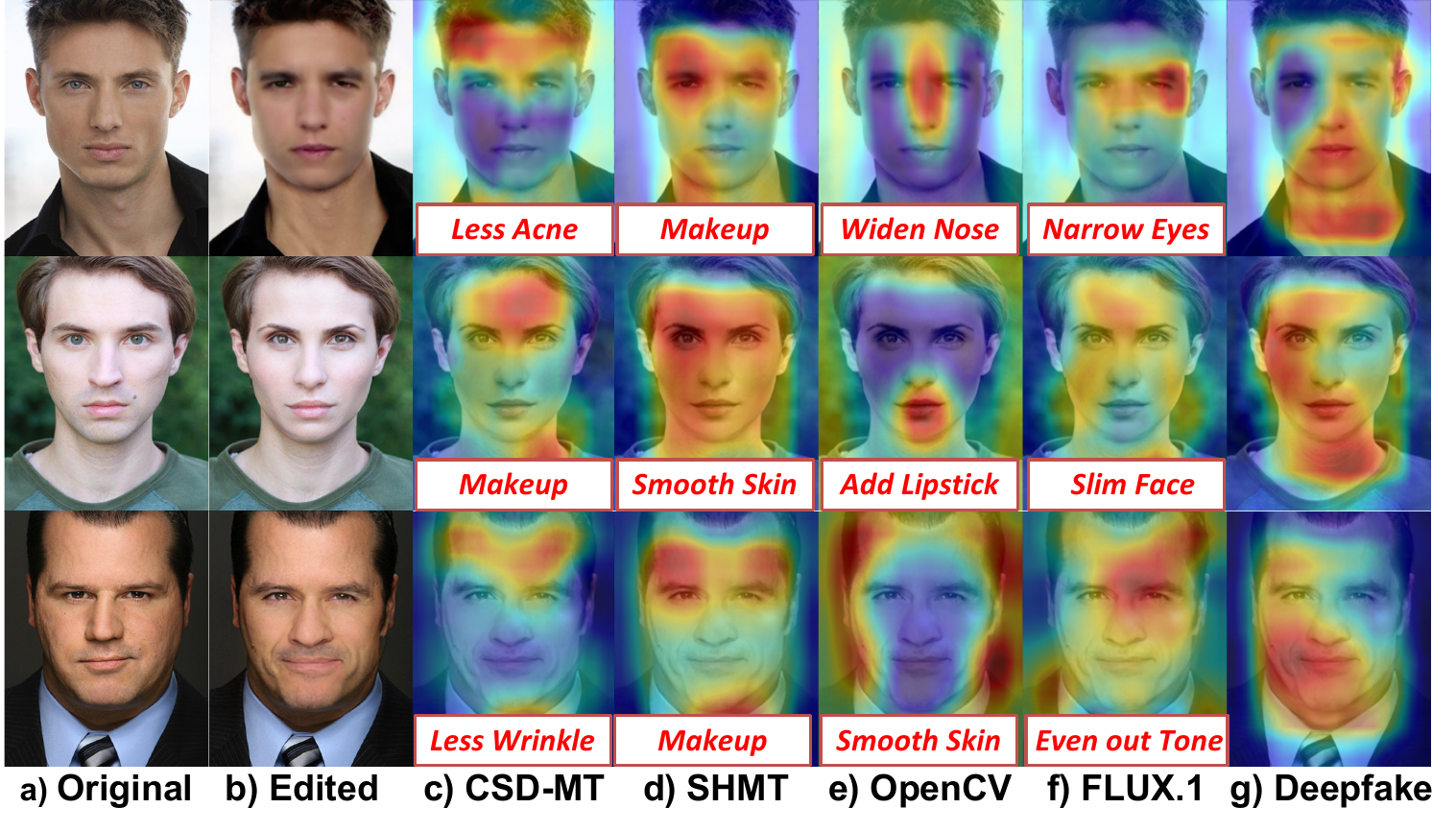}
  \caption{Decisive ROI Visualization for Different Editing Tools. The red-highlighted regions indicate the areas that \Method\ focuses on when attributing each tool. a) Original images; b) Multi-tool edited images; c)-g) Grad-CAM heatmaps for different editing tools. Editing targets are marked with colored boxes for clarity.}
  \label{fig:grad_cam}
\end{figure}

\section{Conclusion}
In this study, we introduced a new task, Multi-tool Image Editing Attribution, which identifies multiple editing tools involved in single-image editing, making the image editing process more transparent and supporting accountability for image authenticity.
We first constructed a 500k-scale dataset, MultiEdit, which covers various combinations based on six editing tools.
Motivated by the analysis of MultiEdit, we proposed \Method, a multi-tool attribution method that consists of two branches that leverage Patchwise Feature Enhancement to effectively learn distinguishable features of individual tools in two domains. Also, we designed an Error-based Curriculum Learning strategy to progressively train the model from easy samples to hard samples to capture the unique and significant tool characteristics in a mixture of editing traces.
Extensive experiments showed the superiority of \Method\ in multi-tool attribution. In the future, we plan to further explore tool-use sequence prediction and test our method on general natural images.

\bibliography{main,supplementary-references}

\clearpage
\appendix
\setcounter{section}{0}
\setcounter{table}{0}
\setcounter{figure}{0}
\setcounter{equation}{0}
\renewcommand\thesection{\Alph{section}}
\renewcommand{\thetable}{S\arabic{table}}
\renewcommand{\thefigure}{S\arabic{figure}}
\renewcommand{\theequation}{S\arabic{equation}}
\renewcommand{\theHsection}{supplementary.\Alph{section}}
\renewcommand{\theHtable}{supplementary.S\arabic{table}}
\renewcommand{\theHfigure}{supplementary.S\arabic{figure}}
\renewcommand{\theHequation}{supplementary.S\arabic{equation}}
\section*{Supplementary Material}
\noindent \textbf{We organize our supplementary material as follows:}
\begin{enumerate}[nosep, leftmargin=*, label={}] 
    \item \textbf{Sec.~\ref{Limitations-and-future-work} Limitations and Future Work}
    \item \textbf{Sec.~\ref{sec:implementation-details} Implementation Details:}
        \begin{itemize}[nosep, leftmargin=*] 
            \item Sec.~\ref{sec:qwen3-eval} Details of using Qwen3-VL as an edit judge
            \item Sec.~\ref{method-design} Details of the \Method\ design
            \item Sec.~\ref{training-process} Details of the training process
        \end{itemize}
    \item \textbf{Sec.~\ref{sec:edited-image-cases} Edited Image Cases}
    \item \textbf{Sec.~\ref{sec:additional-analysis} Additional Analysis:}
        \begin{itemize}[nosep, leftmargin=*]
            \item Sec.~\ref{Hyper-parameter} Key hyperparameter sensitivity
            \item Sec.~\ref{failure-case-analysis} Failure case analysis
            \item Sec.~\ref{sequence-recovery} Attempt at sequence recovery
        \end{itemize}
\end{enumerate}
\vspace{-1em}
\section{Discussion on Limitations and Future Work}
\label{Limitations-and-future-work}
To address the potential attribution issue brought by the recent transition from single-tool image generation or editing using GANs and diffusion to multi-tool image editing, we constructed the infrastructure for multi-tool image editing attribution in this paper by formulating the task, building a large-scale multi-tool-edited image dataset, implementing nine baselines, and providing a new method.
Our work could be regarded as an extension of deepfake detection and attribution to catch up with the recent progress on generative AI, and also a type of sister work of the conventional image provenance analysis that finds images that provide materials for a manipulated one~\cite{8438504}.
Despite this, as a preliminary study, we identify the following limitations:

\textbf{1) The dataset does not consider general-purpose image editing.}
Although the latest image editing tools are capable of general editing, the MultiEdit dataset only considers human face editing.
We focus on face samples because face editing is of high frequency, often related to multiple edits, and has been used to fabricate fake news about celebrities multiple times. This would make our dataset more aligned with real-world cases. We also admit that this could bring some convenience to our analysis due to the constraints of human facial structure. We plan to extend the dataset to edited images in general domains.

\textbf{2) We did not perform tool-use order-level predictions.} We formulate the task as a multi-label classification task regardless of the tool-use order. During our research, we made an initial attempt but finally put this aside as no reasonable results were observed (will show in Sec.~\ref{sequence-recovery}).
We speculate that the effects of each editing tool may partially follow commutativity and thus be indistinguishable in terms of the order. Further experiments on the feasibility of order prediction and its impact on set-level prediction are worth deeper exploration, and we leave it to future work.

\textbf{3) We mainly follow a closed-set setting that limits the attribution models' extensibility.} Following the research line of deepfake attribution~\cite{DNADET,POSE,tmm}, we start with a fundamental closed-set setting, \textit{i.e.,} the attribution model can only attribute the editing of an image to pre-set editing tool candidates. For an image edited by an unseen tool, the model cannot provide a reliable prediction. After validating the attribution possibility through this work, we plan to extend the closed-set setting to an open-set one, making it more applicable to complex real-world scenarios.

\section{Implementation Details}
\label{sec:implementation-details}
\subsection{Qwen3-VL Evaluation Instructions}
\label{sec:qwen3-eval}
To determine whether a specific editing operation has been successfully applied to an image, we use Qwen3-VL to assess the success of the editing, as stated in Section 3.2. The instruction template employed is shown in Figure~\ref{qwen3-VL-effect-eval} below. We further conducted manual sampling verification to assess Qwen3-VL’s competence for this task. Results indicate over 90\% consistency between Qwen3-VL’s judgments and human evaluations, confirming its suitability as a reliable identifier. 

\begin{figure}[htbp]
\vspace{-0.5em}
    \centering
    \begin{tcolorbox}[colframe=black, colback=white]
        \noindent
Now, you are an image editing effect evaluation expert, and please help me complete the following task:  

You will be given:  \\
\quad 1. Original image (Image 1)  \\
\quad 2. Edited image (Image 2)\\  
\quad 3. Description of the editing target. e.g., Add blush to the man, enlarge the eyes of the person, etc.]  
\\
Please compare Image 1 and Image 2, focusing on checking whether the above-mentioned editing operation has been successfully and accurately applied to the image. 
\\
The criteria for your judgment are: whether the changes described in the editing target are clearly and completely presented in Image 2 and conform to the core target of the operation (\textbf{if the editing target is partially implemented, incorrectly implemented, or not implemented at all, it shall be deemed as a failure}).  

PLEASE OUTPUT A SINGLE KEY:VALUE PAIR IN STRICT JSON FORMAT,
DO NOT OUTPUT ANY OTHER INFORMATION!!!!\\
The key is \textbf{\textit{result}}, value is Success or Fail.\\
Two output examples are given below:\\
\{result: \textit{"Success"}\}\\
\{result: \textit{"Fail"}\}
    \end{tcolorbox}
    \vspace{-1em}
    \caption{Prompt template for Qwen3-VL-based image editing effect evaluation}
    \label{qwen3-VL-effect-eval}
    \vspace{-1.5em}
\end{figure}
\subsection{Details of \Method\ Design}
\label{method-design}
\Method\ mainly incorporates two modules: Patchwise feature enhancement~(PFE) and Error-based Curriculum Learning~(ECL) training strategy, shown in Section 4 of the main text. For the PFE module, we apply Slide-Window Cosine Discrete Transform~(SWDCT) to transform the input image from the spatial to frequency domain. And a window size of 16$\times$16 and a step size of 2 were employed in the sliding window process. In the cross attention process, the number of heads is set to 8, and the hidden size in cross attention $d_k$ is set to 768.
For ECL, we set the order $t$ of the polynomial in Equation~(9) to 3 and the age parameter in Equation~(10) is initially set to 0.5, and is incremented by 1 after each training epoch.
\vspace{-1em}
\subsection{Details of Training Process}
\label{training-process}
In all training processes, we use the Accelerate library~\cite{accelerate} for distributed training, and the learning rate is initialized to 3e-4. And we apply a hybrid learning rate scheduling strategy that combines linear warmup and cosine annealing. During the warmup phase (when the current step is less than the warmup steps), the learning rate increases linearly with the number of steps. The scaling factor is calculated as $\frac{(T + 1)}{T_0}$, where $T$ and $T_0$ denote current epoch and warmup epochs, respectively. After the warmup phase, the learning rate decays following a cosine annealing pattern. First, the progress is calculated as the ratio of the steps elapsed since warmup to the total remaining steps after warmup (ranging from 0 to 1). Then, a cosine-based decay curve is generated using the formula $W=0.5*(1+cos\left(\pi * T-T_0)/({T_{all}-T_0})\right)$, where ${T_{all}}$ refers to total training epochs. Finally, the decay factor $W * 0.99 + 0.01$ is applied multiplicatively to the optimizer’s initial learning rates.

For Qwen3-VL and Qwen3-VL-CLS models, we select the Qwen3-VL-30B~\cite{qwen-VL} as the base model and use the instruction in Figure~\ref{qwen3-VL-instruction} for zero-shot inference and training. In the training process of Qwen3-VL-CLS, we apply the low-rank approximation (LoRA)~\cite{LORA} to train both text and image encoders and use binary cross-entropy loss for the multi-label classification loss.

\begin{figure}[htbp]
    \centering
    \begin{tcolorbox}[colframe=black, colback=white]
        \noindent
Now, you are an expert forensic image analysis assistant. \\
For each user image you must assess evidence that any of the following digital image generation or editing tools contributed to the final image: \\
The CANDIDATE TOOLS are Deepfake, GAN, SD, OpenCV, and DiT\\
Definitions: \\
\quad Deepfake=face swap/manipulation pipeline; \\
\quad GAN=generative adversarial network based synthesis/refinement; \\
\quad SD=Stable Diffusion text2image/image2image;\\
\quad OPENCV=classical image processing (filters, geometric, color ops);\\ 
\quad DIT=Diffusion transformer based synthesis/refinement.\\
Decision rule: Evaluate each tool independently; output 1 only if there are concrete visual cues strongly suggesting that tool's involvement (artifacts, blending seams, texture patterns, lighting inconsistencies, diffusion noise fingerprints, classical filter traces). If uncertain, output 0. \\
Output MUST be a single compact JSON object with keys exactly:\\ 
\quad detected-tools (array of strings with positive tools in canonical order);\\
\quad binary (object mapping each tool to 0/1);\\
\quad confidence (object mapping each tool to float in [0,1]);\\ 
\quad reason (short concise English explanation referencing visual cues). "\\
All 5 tools must appear in binary and confidence. Confidence must have at most 3 decimals. NO EXTRA COMMENTARY OUTSIDE JSON!!!!
    \end{tcolorbox}
    \caption{Prompt template for Qwen3-VL-based attribution}
    \label{qwen3-VL-instruction}
\end{figure}
\vspace{-0.5em}
\section{Edited Image Cases}
\label{sec:edited-image-cases}
\begin{figure}[htbp]
    \centering
    \includegraphics[height=0.70\textheight,keepaspectratio]{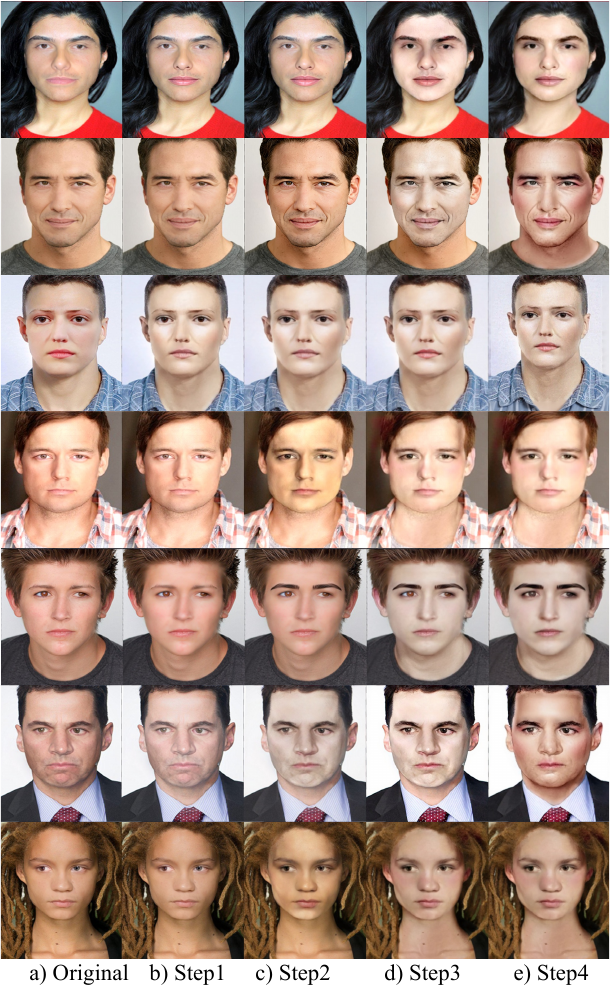}
    \vspace{-1em}
    \caption{Multi-tool edited examples on real faces.}
    \vspace{-1em}
    \label{fig:no_df_editing}
\end{figure}

\begin{figure}[htbp]
    \centering
    \includegraphics[height=0.70\textheight,keepaspectratio]{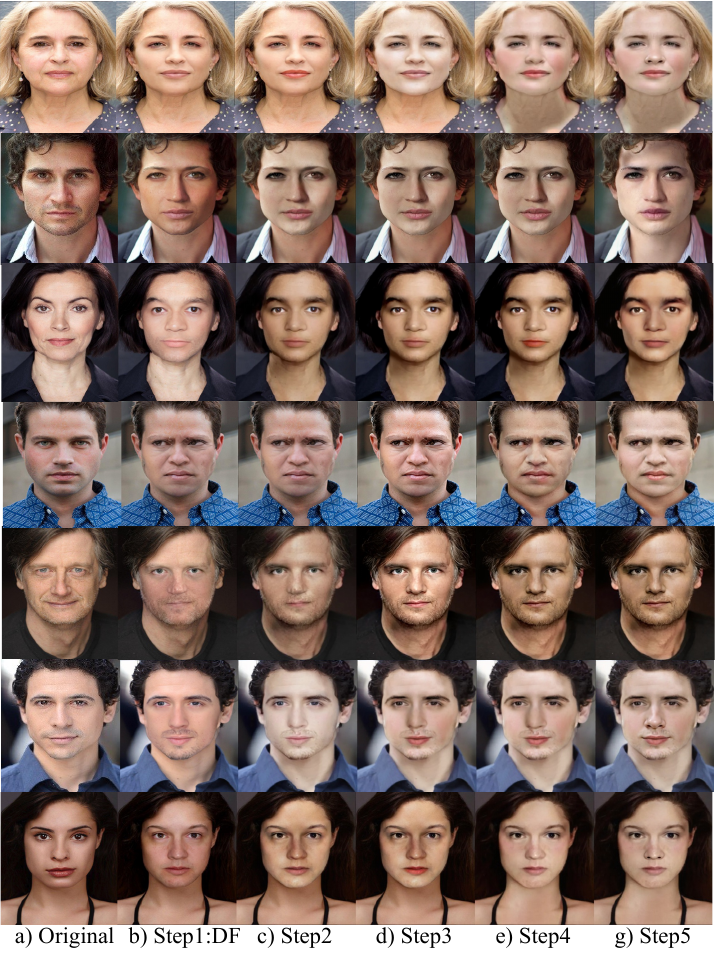}
    \vspace{-1em}
    \caption{Multi-tool edited examples on deepfake faces.}
    \label{fig:with_df_editing}
\end{figure}

Figures~\ref{fig:no_df_editing} and~\ref{fig:with_df_editing} exhibit samples in MultiEdit. During image editing, we select 40 images in total as reference images for the tools CSD-MT and SHMT. More edited samples are provided in the supplementary material.
In some cases, the image edited with multiple tools may feature relatively heavy makeup, resembling stage makeup rather than everyday looks. Because the task evaluation should consider more diverse scenarios, not limited to daily natural looks, we retained these samples for the sake of data diversity.
\vspace{-0.5em}
\section{Additional Analysis}
\label{sec:additional-analysis}
\subsection{Results on Hyperparameter Sensitivity}
\label{Hyper-parameter}
We conducted ablation studies on the key operation and hyperparameters of PFE and ECL, and the results are shown in Table~\ref{tab:ablation_hyper_parameters}. It illustrates that our method exhibits a certain degree of hyperparameter insensitivity. Models deviating from the optimal hyperparameter still maintain high performance on images edited by more than two tools.
\begin{table}[htbp]
    \centering
    \small
    \caption{Ablation on hyperparameters selection in \Method. Slide Window Step = Size $\times$ Ratio. The row in red means the best hyperparameters and corresponding performances.}
  \begin{tabular}{cccccc}
    \toprule
    \multirow{2}{*}{\textbf{Module}} & \multicolumn{2}{c}{\multirow{2}{*}{\textbf{Params.}}}& \multicolumn{3}{c}{\textbf{\# Editing Steps}} \\
    \cmidrule(lr){4-6}
    & & & 3 & 4 & 5 \\
    \hline
    \multirow{12}{*}{SWDCT} & \multirow{12}{*}{Size;Ratio}
    & 8$\times$8;1/8 & 0.8832 & 0.7832 & 0.7256 \\
    & & 8$\times$8;1/4 & 0.8798 & 0.7796 & 0.7185 \\
    & & 8$\times$8;1/2 & 0.8837 & 0.8012 & 0.7115 \\
    & & \cellcolor{pink!50}16$\times$16;1/8 & \cellcolor{pink!50}0.9309 & \cellcolor{pink!50}0.8918 & \cellcolor{pink!50}0.8740 \\
    & & 16$\times$16;1/4 & 0.9387 & 0.8848 & 0.8675 \\
    & & 16$\times$16;1/2 & 0.9247 & 0.8564 & 0.8703 \\
    & & 32$\times$32;1/8 & 0.9198 & 0.8756 & 0.8564 \\
    & & 32$\times$32;1/4 & 0.9227 & 0.8530 & 0.8501 \\
    & & 32$\times$32;1/2 & 0.9017 & 0.8534 & 0.8567 \\
    & & 64$\times$64;1/8 & 0.8917 & 0.8321 & 0.8032 \\
    & & 64$\times$64;1/4 & 0.8546 & 0.8062 & 0.8042 \\
    & & 64$\times$64;1/2 & 0.8347 & 0.7865 & 0.7564 \\
    \hline
    \multirow{5}{*}{ECL} & \multirow{5}{*}{\makecell{Increase\\ rate of $\lambda$}} 
    & 0.5 & 0.9423 & 0.8716 & 0.8392 \\
    & & \cellcolor{pink!50}1.0 & \cellcolor{pink!50}0.9309 & \cellcolor{pink!50}0.8918 & \cellcolor{pink!50}0.8740 \\
    & & 1.5 & 0.9323 & 0.8716 & 0.8692 \\
    & & 2.0 & 0.9218 & 0.8567 & 0.8548 \\
    & & 2.5 & 0.9134 & 0.8874 & 0.8632 \\
    \bottomrule
  \end{tabular}
  \label{tab:ablation_hyper_parameters}
\end{table}
\subsection{Failure Case Analysis}
\label{failure-case-analysis}
\begin{figure}[htbp]
    \centering
    \includegraphics[width=0.90\linewidth]{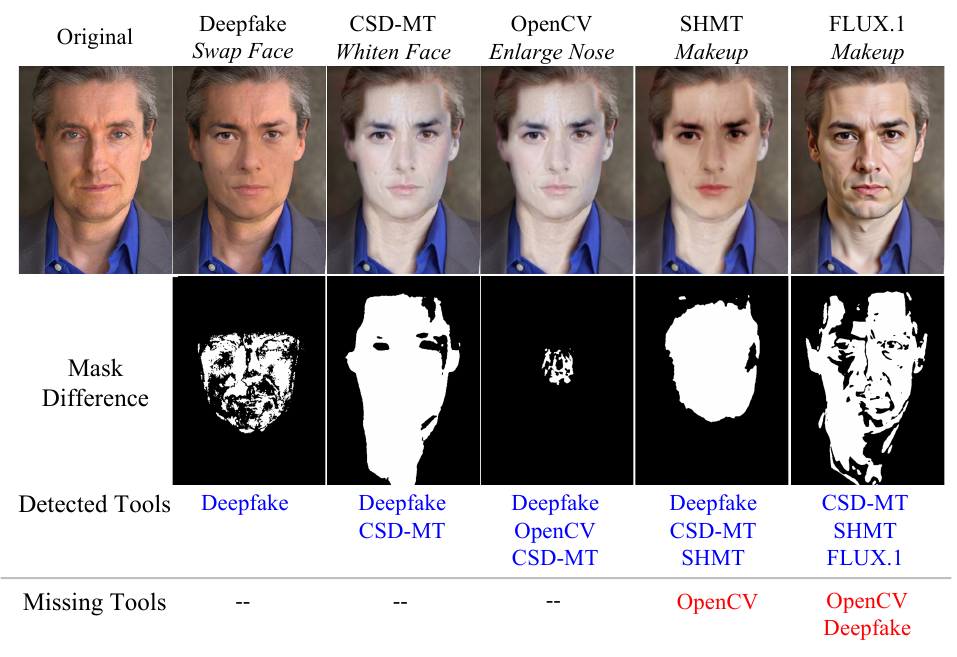}
    \caption{A failure case where some tools are in editing process but not detected. The missed tools are in \textcolor{red}{red}.}
    \vspace{-2em}
    \label{failure-cases}
\end{figure}

Figure~\ref{failure-cases} depicts a group of sequentially edited images that DPEC failed to detect some of the editing tools. In this sequential editing process, the traces of the preceding small-area editing tools are obfuscated by later large-area edits, resulting in missed detection on preceding editing tools. This is somewhat aligned with the results in Table 6 of the main text that the size of the pixel-level differences between two consecutive edited images would influence the detectability of the traces left by previous tools. 
\subsection{An Attempt at Tool-Use Sequence Recovery}
\label{sequence-recovery}
As we stated in the main text, we formulate multi-tool image attribution as a multi-label classification task regardless of the tool-use order.
That is, we perform a \textit{set-level} attribution rather than a \textit{sequence-level} one.
In this section, we made a preliminary attempt at sequence-level attribution (\textit{i.e.}, tool-use sequence recovery for an edited image) to know whether the editing order is predictable (or whether the tool-use does not hold commutativity).
Unlike previous work modeling this task as an autoregressive prediction task, since images are exhibited as static media and thus non-sequential inherently,  we change the output of the model \Method\ to
\begin{equation}
   M= 
   \begin{bmatrix} 
        [0 & 0 & \cdots & 1 & 0] \\ 
        \vdots & \vdots & \ddots & \vdots & \vdots \\
        [1 & 0 & \cdots & 0 & 0] 
   \end{bmatrix}_{N \times T},
\end{equation}
where $N$ denotes the maximum editing steps and $T$ denotes the total number of used editing tools. $M[i,j]=1$ means that the model predicts $Tool_j$ is used at the $i$-th editing step.

For evaluation, we apply a sequence-level exact match score (EM) that requires both the tool set and order to be correct. The model achieves 1.67\% in EM on the images edited by 5 tools and 12.64\% in EM on average among all test samples. These results show that the tool-use sequence recovery is difficult for the current model, and further exploration is expected.

\end{document}